\documentclass{article}

\usepackage{PRIMEarxiv}

\usepackage[utf8]{inputenc} 
\usepackage[T1]{fontenc}    
\usepackage{hyperref}       
\usepackage{url}            
\usepackage{booktabs}       
\usepackage{amsfonts}       
\usepackage{nicefrac}       
\usepackage{microtype}      
\usepackage{lipsum}
\usepackage{fancyhdr}       
\usepackage{graphicx}       
\usepackage{hyperref}       
\usepackage{url}            
\usepackage{booktabs}       
\usepackage{amsfonts}       
\usepackage{nicefrac}       
\usepackage{microtype}      
\usepackage{xcolor}         
\usepackage{natbib}
\usepackage{graphicx} 
\usepackage{booktabs}
\usepackage[table]{xcolor}
\usepackage{amsmath}
\usepackage{colortbl}
\definecolor{mygray}{gray}{0.8}
\usepackage[most]{tcolorbox}
\usepackage{wrapfig}
\usepackage{ulem}
\usepackage{subcaption}
\usepackage{extarrows}
\usepackage{multirow}
\usepackage{makecell}

\definecolor{blockblue}{RGB}{214,234,248}
\definecolor{blockgreen}{RGB}{221,237,211}
\definecolor{blockorange}{RGB}{254,235,201}
\definecolor{blockpink}{RGB}{251,228,236}

\graphicspath{{media/}}     

\title{Generalizable Geometric Image Caption Synthesis}

\author{
  Author1, Author2 \\
  Affiliation \\
  Univ \\
  City\\
  \texttt{\{Author1, Author2\}email@email} \\
   \And
  Author3 \\
  Affiliation \\
  Univ \\
  City\\
  \texttt{email@email} \\
}

\author{
 \textbf{Yue Xin\textsuperscript{1,2*$\dagger$}},\quad
 \textbf{Wenyuan Wang\textsuperscript{1,3*$\dagger$}},\quad
 \textbf{Rui Pan\textsuperscript{1*}},\\
  \textbf{Ruida Wang\textsuperscript{1}},\quad
  \textbf{Howard Meng\textsuperscript{1}},\quad
  \textbf{Renjie Pi\textsuperscript{4}},\quad
  \textbf{Shizhe Diao\textsuperscript{4}},\quad
  \textbf{Tong Zhang\textsuperscript{1}}\\
 \textsuperscript{1}University of Illinois Urbana-Champaign (UIUC), \\
 \textsuperscript{2}Shanghai Jiao Tong University, \quad \textsuperscript{3}Rutgers University, \quad
 \textsuperscript{4}NVIDIA\\ \\
 \href{https://machinephoenix.github.io/GeoReasoning_blog/}{Project Page} \quad \href{https://huggingface.co/datasets/ScaleMath/GeoReasoning}{Dataset} \quad \href{https://github.com/MachinePhoenix/GeoReasoning}{Code}
}

\begin{document}
\maketitle

\def\thefootnote{*}\footnotetext{Core contributors.}
\def\thefootnote{$\dagger$}\footnotetext{Work done during a research internship at UIUC.}

\begin{abstract}
Multimodal large language models have various practical applications that demand strong reasoning abilities. Despite recent advancements, these models still struggle to solve complex geometric problems. A key challenge stems from the lack of high-quality image-text pair datasets for understanding geometric images. Furthermore, most template-based data synthesis pipelines typically fail to generalize to questions beyond their predefined templates. In this paper, we bridge this gap by introducing a complementary process of Reinforcement Learning with Verifiable Rewards (RLVR) into the data generation pipeline. By adopting RLVR to refine captions for geometric images synthesized from 50 basic geometric relations and using reward signals derived from mathematical problem-solving tasks, our pipeline successfully captures the key features of geometry problem-solving. This enables better task generalization and yields non-trivial improvements. Furthermore, even in out-of-distribution scenarios, the generated dataset enhances the general reasoning capabilities of multimodal large language models, yielding accuracy improvements of 2.8\%–4.8\% in statistics, arithmetic, algebraic, and numerical tasks with non-geometric input images of MathVista and MathVerse, along with 2.4\%-3.9\% improvements in Art \& Design and Tech \& Engineering tasks in MMMU.
\end{abstract}

\section{Introduction}
Multimodal Large Language Models (MLLMs) have exhibited impressive capabilities across a variety of vision-related tasks, including Visual Question Answering (VQA), visual grounding, and image captioning. Recent MLLMs, such as Qwen2.5-VL, Intern2.5-VL, and LLaVA-Next~\citep{bai2025qwen2-5vl,chen2024internvl,liu2024llavanext}, have shown superior performance compared to specialized vision models across a wide range of visual tasks, highlighting the potential of unified multimodal architectures. As the field advances, there has been increasing interest in enhancing the reasoning capabilities of MLLMs~\citep{jaech2024openai,shao2024visual}, which is regarded as a crucial factor in extending the performance boundaries of these models. Among various reasoning tasks, mathematical reasoning~\citep{zhang2024mathverse} has attributed particular attention due to its structured problem-solving nature, offering a clear pathway for MLLMs to develop and improve their reasoning skills.

Research from MathVerse~\citep{zhang2024mathverse} indicates that MLLMs perform best when the input is purely textual, while their performance declines significantly with visual-only inputs. This highlights the urgent need for MLLMs to develop strong cross-modal reasoning capabilities, which involves accurately and comprehensively transferring information from the image to the text. Although numerous geometry and math datasets have been introduced~\citep{lu2023mathvista, zhang2024mavis, wang2024measuring} to boost various aspects of model performance, high‑quality datasets explicitly designed for cross‑modal reasoning remain scarce. That is primarily because in existing datasets, the alignment between images and captions is often asymmetrical. For instance, in geometric problems, two lines of equal length can be easily described textually but may not be correspondingly annotated or visually distinct in the image. Such discrepancies hinder the model's ability to learn robust cross-modal reasoning.

Meanwhile, Reinforcement Learning (RL) has been demonstrated to significantly improve model reasoning and generalization capabilities~\citep{guo2025deepseek}. Its reward-driven framework is particularly effective for cross-modal reasoning, allowing models to optimize decision-making through interactive feedback~\citep{deng2025openvlthinker,peng2025lmm,huang2025vision}. Building on these insights, we employ the RAFT method~\citep{Dong2023RAFT} and design a reward function that incorporates both reasoning and caption rewards. This facilitates the alternating optimization of dataset quality and model reasoning abilities, leading to improved results on complex multimodal tasks.

To bridge the gap between visual and linguistic modalities, we propose an RL-based data refinement engine that iteratively enhances data quality. Utilizing this pipeline, we introduce a novel geometry dataset comprising 10,000 image-caption pairs. To the best of our knowledge, this is the first high-quality dataset in which visual and textual information are fully aligned and generalize well to out-of-distribution tasks, making it a valuable resource for improving cross-modal reasoning. Experimental results demonstrate that our dataset significantly enhances models' cross-modal reasoning abilities and their performance on geometric image textualization tasks. Furthermore, models trained on our dataset exhibit strong generalization capabilities on other mathematics-focused benchmarks, including MathVerse and MathVista.
In summary, our main contributions are summarized as follows:

\begin{itemize}
    \item We introduce \textbf{GeoReasoning-10K}, a new dataset containing 10,000 carefully constructed image-caption pairs where visual and textual information are fully equivalent. This dataset serves as a high-quality resource for training cross-modal reasoning models.

    \item We propose \textbf{Geo-Image-Textualization}, a scalable RL-based framework for generating and refining high-quality synthetic image-caption pairs in geometry. Our iterative RL-driven optimization significantly enhances data alignment and semantic accuracy, and demonstrates generalization to out-of-domain geometric tasks.

    \item Extensive experiments show that the improvements facilitated by GeoReasoning extend beyond geometric tasks to non-geometric mathematical tasks and even to non-mathematical domains such as art and engineering.
\end{itemize}

\begin{figure}[!t]
    \centering
    \includegraphics[width=0.9\linewidth]{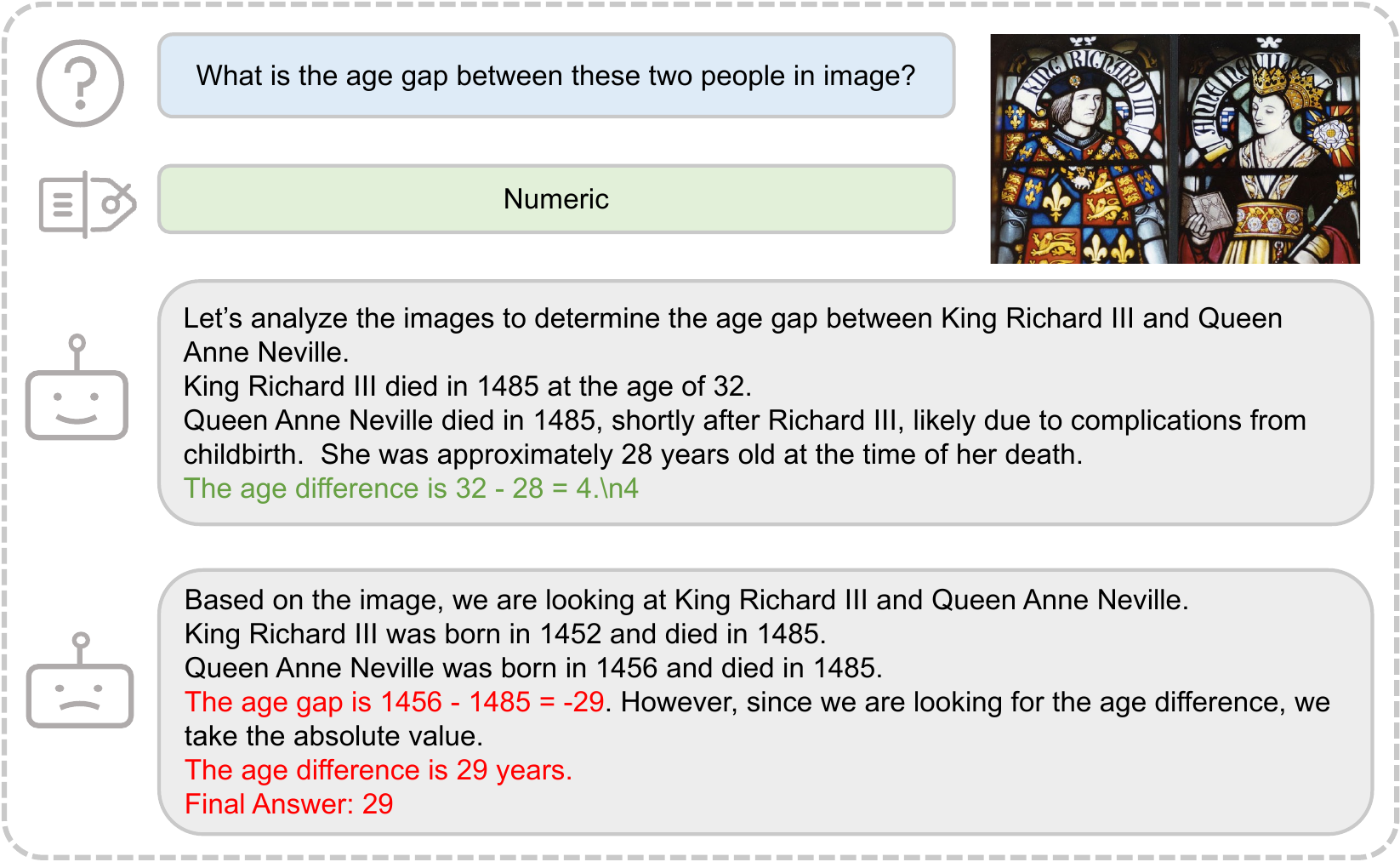}
    \caption{\textbf{Examples of generalization}. MLLMs learn from our synthetic geometric mathematical problems and generalize to algebraic cases with even non-geometric input images.}
    \label{case_algebraic}
\end{figure}

\section{Related Works}

\subsection{Data Generation}
Recent studies have highlighted the scarcity of high-quality geometry image–caption datasets, which limits fine‑grained cross‑modal reasoning in geometric tasks. Following AlphaGeometry~\citep{trinh2024solving}, Autogeo~\citep{huang2025autogeo} proposed an automatic generation engine to produce image-caption pairs, constructing a 100K dataset named AutoGeo-100k. MATHGLANCE~\citep{sun2025mathglance} introduced GeoPeP, a perception-oriented dataset of 200K structured geometry image-text pairs explicitly annotated with geometric primitives and spatial relationships. MagicGeo~\citep{wang2025magicgeo} formulates diagram synthesis as a coordinate optimization problem, ensuring formal geometric correctness via solvers before coordinate‑aware text generation.


Despite the advances, existing pipelines still struggle to guarantee full modality alignment, i.e., captions frequently omit visual details, while images lack exhaustively aligned textual descriptions. 


\subsection{Image Captioning}
Image captioning aims to generate comprehensive descriptions of visual content and has been widely studied for natural images. While general-purpose multimodal large language models (MLLMs) such as mPLUG-Owl2~\citep{ye2024mplug}, MiniGPT-4~\citep{zhu2023minigpt}, and BLIP-3~\citep{xue2024xgen} can perform captioning to some extent, their effectiveness is often limited by insufficient fine-grained cross-modal alignment. Image-Textualization~\citep{pi2024image} mitigates this issue by integrating multiple vision experts to produce more detailed and accurate captions.

However, the potential of utilizing image captioning to enhance geometric reasoning capacity remains underexplored. OmniCaptioner~\citep{lu2025omnicaptioner} proposes a unified visual captioning framework that converts diverse images into fine‑grained textual descriptions. Nonetheless, its geometric annotations are derived from AutoGeo and MAVIS, largely relying on synthetic or loosely aligned pairs rather than fully equivalent visual‑textual representations. Moreover, the scarcity of high-quality geometric image-caption pairs makes it difficult to accurately extract and align geometric information. As a result, current models underperform on geometric image textualization compared to natural image captioning and general visual reasoning benchmarks.

\section{Methods}
In this section, we introduce our Geo-Image-Textualization data generation pipeline first, followed by our RAFT method used for data refinement.

\subsection{Geo-Image-Textualization Data Generation Engine}
The proposed data generation pipeline mainly contains three parts: the relation sampling, image-caption pair generation, and question-answer generation procedure, as shown in Figure.~\ref{fig:geometry-data-synthesis-pipeline}.

\begin{figure}[!t]
  \centering
  \includegraphics[width=0.27\textwidth]{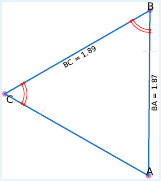}
  \label{fig:ex1}
  \hfill
  \includegraphics[width=0.31\textwidth]{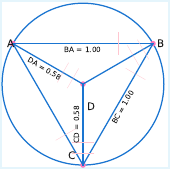}
  \label{fig:ex2}
  \hfill
  \includegraphics[width=0.28\textwidth]{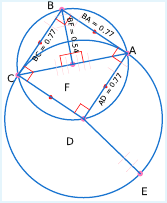}
  \label{fig:ex3}
  \caption{\textbf{Symbolically synthesized geometric images}. These geometry problems are symbolically composed from our relation library, corresponding to easy, medium, and hard difficulty levels, respectively, where the pink ticks and red arcs indicate equal-length segments and equal angles. The symbolic engine can generate images with infinite types and difficulties. For visual clarity, this figure has a fixed set of colors, font sizes, and line thicknesses compared to the original images in our constructed dataset. Please refer to the original dataset for precise details.}
\end{figure}

\subsubsection{Relation Sampling} 
We develop the Geo-Image-Textualization pipeline upon the data generation procedure in AlphaGeometry. In our framework, \textit{Relations} act as fundamental construction operations that systematically generate diverse yet semantically coherent geometric premises for subsequent theorem synthesis. Each relation (e.g., \texttt{angle\_mirror}, \texttt{circumcenter}, etc.) encodes a precise geometric procedure---such as reflecting a point across an angle bisector or locating the circumcenter of a triangle. In addition to the construction rule, each definition maintains dependency metadata, specifying which primitive objects (points, lines, circles) and prior constructions it depends on. This enables the symbolic engine to get the minimal set of premises required to derive a given theorem. 



After sampling relations, each relation is converted into a clause, with associated point variables. For instance, the relation \texttt{angle\_mirror x a b c} denotes: given points \texttt{a}, \texttt{b}, and \texttt{c}, construct point \texttt{x} as the reflection of \texttt{c} across angle~$\angle ABC$. Finally, the system constructs a graph-based representation in AutoGeo~\citep{huang2025autogeo} to model geometric problems.  Each clause, corresponding to either a geometric construction or a relational assertion, is incorporated into the graph by instantiating nodes for geometric entities (e.g., points, lines, circles) and establishing edges that encode their interdependencies. Before adding each clause, the system verifies the logical correctness of the selected set of predefined geometric rules.




\begin{figure}[t]
  \centering
  \includegraphics[width=0.95\textwidth]{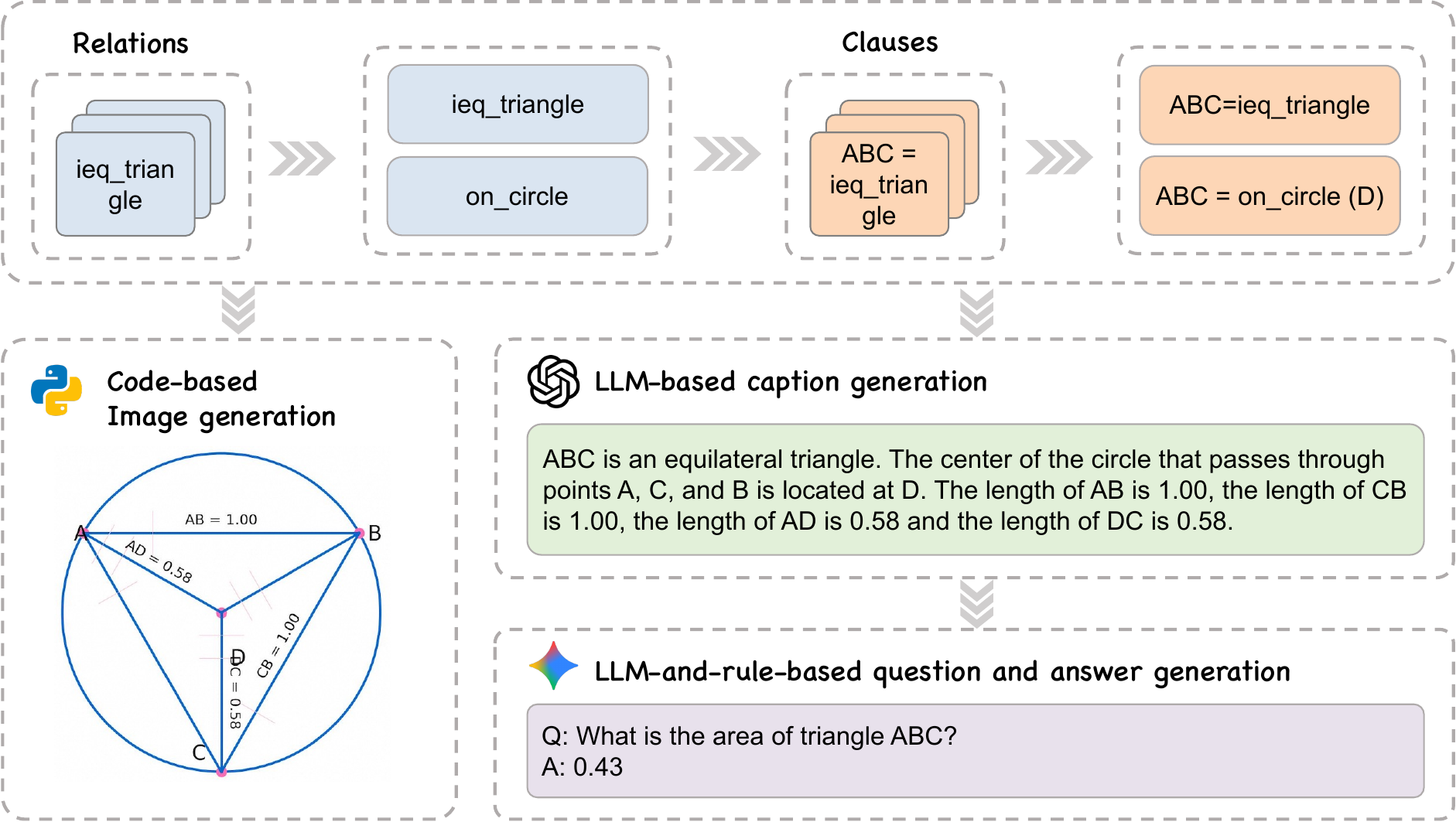}
  \caption{\textbf{The geometry data synthesis pipeline}, where a graph-based representation similar to AutoGeo~\citep{huang2025autogeo} is employed for generating the final geometry images. The relation library comprises over 50 basic geometric relationships that can be composed into complex ones, providing comprehensive coverage for geometric problems of various difficulties. The image-caption pair is utilized for the SFT stage, while the caption-QA pair is for the RLVR stage.}
  \label{fig:geometry-data-synthesis-pipeline}
\end{figure}

\subsubsection{Image-Caption Pair Generation}


The geometric graph encodes all relevant entities, including points, lines, and circles, enabling the straightforward rendering of basic geometric elements, similar to AutoGeo. However, a fundamental limitation of AutoGeo is that the captions cannot be directly inferred from the rendered images because the visual content and the textual description are not semantically aligned. We argue that this misalignment constitutes a critical bottleneck in cross-modal reasoning.

To address this issue, we introduce a set of visual augmentation strategies that explicitly encode semantic relationships within the image, mainly including the following properties, as shown in Figure~\ref{fig:geometry-data-synthesis-pipeline}.
\begin{enumerate}
    \item \textbf{Segment Equality Representation:} We use short perpendicular ticks to indicate equal-length line segments. When multiple pairs of equal segments exist, we distinguish them using a different number of ticks (e.g., one tick, two ticks).

    \item \textbf{Angle Annotations:} For angles that are integer multiples of 15° within the range $[15^\circ, 165^\circ]$, we directly annotate the angle values within the image.

    \item \textbf{Parallel and Perpendicular Indicators:} Parallel lines are marked using matching directional triangles, and right angles are indicated using a small square symbol at the vertex.

    \item \textbf{Equal Angle Representation:} Equal angles are denoted by marking them with the same number of arcs, consistent with conventional geometric notation.

    \item \textbf{Intersection and Collinearity:} Dashed lines are used to explicitly indicate intersections and collinearity relationships among points or segments.
\end{enumerate}



For each clause in the symbolic representation, we apply a refined, rule-based template to convert it into natural language. These captions accurately describe the geometric diagram, including object relationships, special angle values, and other visual properties. Additionally, the captions explicitly state the lengths of specific line segments when such information is visually annotated in the image. By ensuring that all semantic content present in the image is mirrored in the caption, we achieve full cross-modal alignment. 

Building upon these strategies, we construct a comprehensive image-caption generation engine. The engine is entirely rule-based, offering a fast, rigorous, reliable, and cost-effective solution for producing high-quality image-caption pairs to serve as a solid foundation for downstream multi-modal reasoning tasks.

\subsection{Question-Answer Pair Generation}
The most fundamental requirements for generating questions lie in three aspects. First, the generated question should be based on the caption, i.e., should not be irrelevant to the caption. Second, any information already present in the caption should be removed, as this would dilute the impact of the caption and make the evaluation of caption quality harder. Last, the question should be compatible with the given information, so that it can be logically answered based on what is provided.

Based on these requirements, we propose a rule-and-LLM-based pipeline to systematically generate the question and answer based on the pre-generated caption. Specifically, we prompt the large model (Gemini 2.5 Flash) with rubric-based instructions to generate initial questions based on the caption, while also letting the model flag those questions inconsistent with the caption. For the inconsistent questions, we then switch to a different prompt, encouraging the model to incorporate additional information and formulate new questions accordingly. This process continues until a self-consistent question is generated for the first time. The detailed two-stage prompt design is provided in Appendix~\ref{app:qa}.

\subsection{RLVR Framework for Data Refinement}
\begin{figure}[!t]
    \centering
    \includegraphics[width=0.95\linewidth]{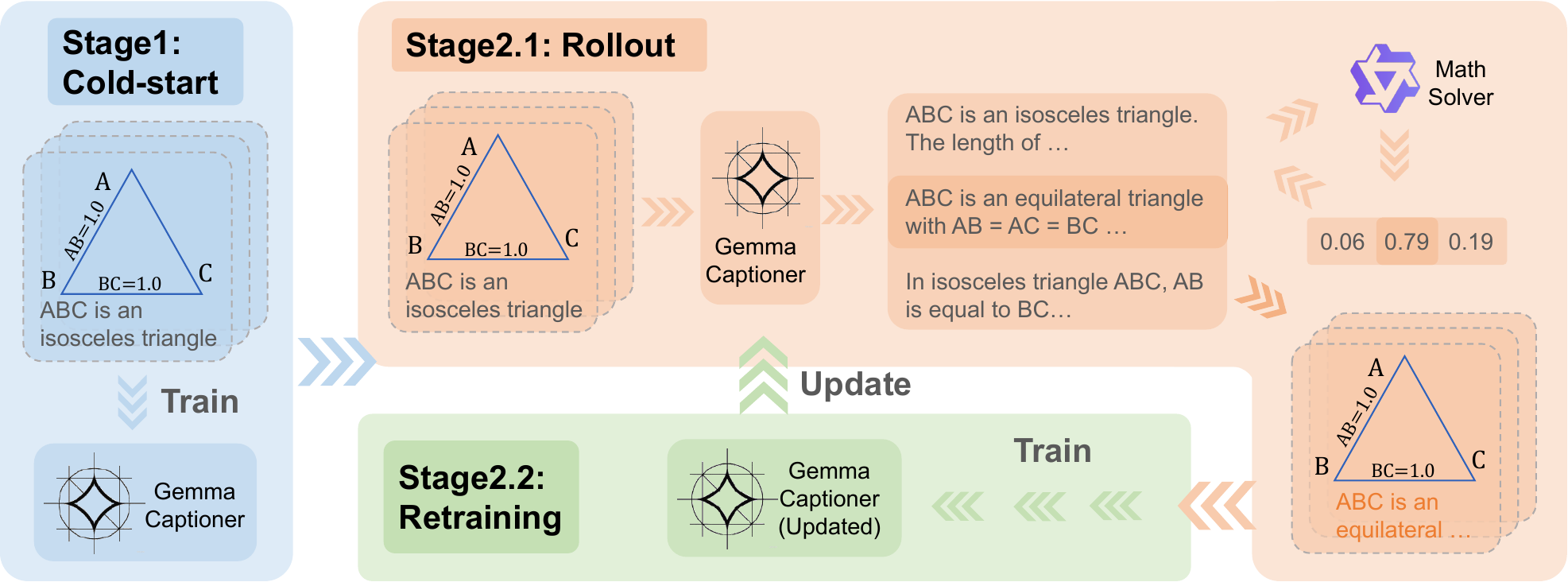}
    \caption{\textbf{The RLVR training framework}. In Stage 1, the model is trained to develop a preliminary ability to generate image captions. In Stage 2, an alternating optimization strategy is employed to jointly refine the generated captions and enhance the model's overall performance. The data of Stage 1 comes from the rule-based image-caption generation pipeline illustrated in Figure~\ref{fig:geometry-data-synthesis-pipeline}.}
    \label{raft}
\end{figure}

Our proposed RLVR framework iteratively optimizes both the model and the dataset through a novel alternating paradigm. The method consists of two phases: (1) a cold-start supervised fine-tuning phase to bootstrap initial captioning capabilities, and (2) an RLVR phase with RAFT~\citep{Dong2023RAFT} that cyclically refines the dataset and model via reinforcement learning. The overall framework is shown in Figure~\ref{raft}.

\subsubsection{Cold-Start Phase}
To initialize the model’s ability to generate geometrically aligned captions, we first perform supervised fine-tuning (SFT) on the base vision language model using the GeoReasoning-10K dataset. This phase minimizes the standard cross-entropy loss:
\begin{align}
    \mathcal{L}_{\text{SFT}} = -\mathbb{E}_{(I, c^\star)\sim \mathcal{D_0}}[\log P_{\theta_0}(c|I)]
\end{align}
where $I$ denotes an input geometric image, $c^\star$ and $c$ indicate the ground-truth caption and the predicted caption, respectively, and $D_0$ represents the initial dataset. The model parameter $\theta_0$ is optimized to establish basic image-to-text mapping capabilities.

\subsubsection{RLVR Phase}
The RLVR phase with RAFT operates in alternating stages, as shown in Figure~\ref{raft}.

\paragraph{Rollout Experience Generation}
Suppose the current iteration is $t$. For each image $I$ in the current dataset $\mathcal{D}_t$, we first generate $N$ candidate captions $\{c_i\} (i=1, 2,\cdots, N)$ using the current vision language model with parameter $\theta_{t}$. Then, we utilize a specifically designed reward function $R(c_i, Q_i, c^\star)$ (detailed introduced in Section~\ref{sec-reward}) to score each caption. Last, we retain the top-K caption $c_{\text{best}}=\arg \max_{c_i} R(c_i, Q_i, c^\star)$ to update the current dataset and construct the refined dataset $\mathcal{D}_{t+1}$.

\paragraph{Model retraining}
We update the model by training on $\mathcal{D}_{t+1}$ for one epoch, which is:
\begin{align}
    \theta_{t+1} = \arg \max_{\theta_t} \mathbb{E}_{(I, c_{\text{best}})\sim \mathcal{D}_{t+1}}[\log P_{\theta_t}(c_{\text{best}}|I)]
\end{align}

This iterative process continues for $T=5$ epochs, progressively enhancing both dataset quality and model performance.

\subsubsection{Reward function}
\label{sec-reward}
The composite reward $R(c, q, c^\star)$ balances task correctness and caption-image alignment, as shown in Eq.~\ref{reward}:
\begin{align}
    \label{reward}
    R(c, I) = \lambda_r\cdot R_{\text{reasoning}}(c, q) + (1-\lambda_r)\cdot R_{\text{caption}}(c, c^\star)
\end{align}

\paragraph{Reasoning reaward}
To evaluate a candidate caption’s utility for solving downstream tasks, we leverage a frozen large language model of Qwen2.5-7B-Instruct~\citep{yang2024qwen2-5} to generate an answer $a \sim P_{\text{LLM}}(a|q, a^\star, c)$ where $q, a^\star$ is the geometric question and its groundtruth answer generated by a reasoning model (Gemini2 .5 Flash) corresponding to the caption $c^\star$ in advance. As encouraged by mainstream RL process, we check both the format and correctness of the answer, which is:
\begin{align}
    R_{\text{Reasoning}} = s_{c}\cdot \mathbb{I}(a=a^\star) + (1-s_{c}) \cdot \mathbb{F}(a)
\end{align}
where $\mathbb{F}(\cdot)$ denotes the format checking function, and $s_c$ indicate the weight of correctness, set as $0.9$ in the experiments.

\paragraph{Caption reward}
To prevent reward sparsity during early training, we measure semantic relevance between $c$ and the ground-truth caption $c^\star$ using ROUGE and BLEU-4, as shown in Eq.~\ref{caption_reward}:
\begin{align}
    \label{caption_reward}
    R_{\text{caption}} = w_r \cdot ROUGE(c, c^\star) + (1-w_r) \cdot BLEU(c, c^\star)
\end{align}
where $w_r$ represents the weight of ROUGE score, set as $0.7$ in the experiments.

\begin{figure}[!t]
    \centering
    \includegraphics[width=1.0\linewidth]{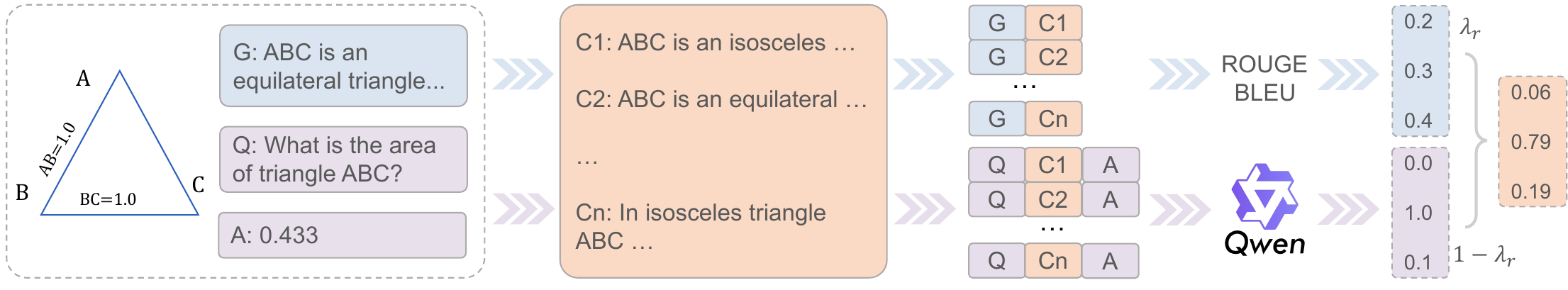}
    \caption{\textbf{The reward function}. Given the Generated caption (G), Question (Q),  and Answer (A), the reward function measures the caption's quality from two aspects: 1) its reasoning reward, and 2) caption reward, as forumated as $R(c, I) = \lambda_r\cdot R_{\text{reasoning}}(c, q) + (1-\lambda_r)\cdot R_{\text{caption}}(c, c^\star)$. \textbf{Reasoning reward} stands for the caption's relevance to the image and question, especially the ability to capture the key reasoning information for solving the question. \textbf{Caption reward} is an auxiliary reward signal that measures the caption's similarity to the ground truth caption.}
    \label{raft}
\end{figure}

\section{Experiments}
\subsection{Experimental Setup}
\label{exp-0}

Our experiments utilize Gemma3-4B~\citep{gemma3}, a commonly used lightweight multimodal architecture with strong mathematical reasoning capabilities, as our base model. All in-domain experiments are conducted on MathVista~\citep{lu2023mathvista} and MathVerse~\citep{zhang2024mathverse}, two established mathematical reasoning benchmarks focusing on visual and mathematical problem-solving.

\subsection{In-Domain Performance of GeoReasoning-10K}
\label{exp-2}
In this section, we verify the effectiveness and scalability of our proposed dataset on in-domain benchmarks. Several commonly-used datasets in this field are chosen as baselines, including AutoGeo~\citep{huang2025autogeo}, GeoPeP~\citep{sun2025mathglance}, GeoGPT4V~\citep{geogpt4}, Geo170K~\citep{geo170k}, GeoQA~\citep{geoqa}, and MathVision~\cite{mathvision}.

It can be observed from Table~\ref{tab3:compare} that the model trained on GeoReasoning-10K obtains better mathematical reasoning performance compared to that trained on other caption datasets. This improvement mainly concentrates on in-domain mathematical subtasks, such as geometry, algebra, science, statistics, and most subtasks in MathVerse. The performance gain can be attributed to the symbolic synthesis process of our pipeline, which allows an infinite number of possible geometry problem types and offers diverse difficulty levels for the generated images.

Besides, the performance of models trained on the full datasets on MathVista and MathVerse is shown in Table~\ref{tab3:sizes} in Appendix~\ref{app:sizes}.

\begin{table*}[!h]
    \centering
    \small
    
    \caption{\textbf{Better In-Domain Performance}. Accuracy of Gemma3-4B models trained on 10k random samples of each dataset over 4 trials. Results on several subtasks in MathVerse and MathVista are also reported here.}\renewcommand{\arraystretch}{1.2}
    \setlength{\tabcolsep}{2.5pt}
    \resizebox{\textwidth}{!}{%
    \begin{tabular}{@{}llllll|lllllllll@{}}
        \toprule
        & \multicolumn{5}{c|}{MathVista ($\uparrow$)} & \multicolumn{4}{c}{MathVerse ($\uparrow$)} \\
        \cmidrule(lr){2-6} \cmidrule(lr){7-10} 
        & \multicolumn{1}{c}{Overall} &  \multicolumn{1}{c}{Geometry} & \multicolumn{1}{c}{Algebra} & \multicolumn{1}{c}{Science} & \multicolumn{1}{c|}{Statistic} & \multicolumn{1}{c}{Overall} & \makecell{Vision\\-Dominant} & \makecell{Text\\-Dominant} & \makecell{Text\\-Lite} \\
        \midrule
        Base & 46.2 & 60.7 & 59.1 & 53.3 & 43.2 & 25.2 & 24.0 & 32.0 & 25.9\\
        AutoGeo & 47.8$\pm$0.8 & 62.3$\pm$2.4 & 60.2$\pm$1.9 & 52.5$\pm$1.2 & 44.1$\pm$0.9 & 24.6$\pm$0.4 & 22.3$\pm$1.4 &35.2$\pm$0.7 &26.7$\pm$1.3\\
        GeoPeP & 47.5$\pm$0.4 & 61.0$\pm$2.3 & 59.6$\pm$1.8 & 54.1$\pm$0.6 & 44.2$\pm$0.8 & 24.2$\pm$0.2 & 21.7$\pm$0.9 & 33.7$\pm$0.3 & 25.7$\pm$1.3\\
        GeoGPT4 & 47.5$\pm$0.2 & 60.5$\pm$0.7 & 59.3$\pm$1.3 & 54.1$\pm$1.5 & 44.6$\pm$1.0 & 25.2$\pm$0.5 & 22.4$\pm$0.8 & 36.4$\pm$1.4 & 26.9$\pm$1.0\\
        Geo170K & 47.6$\pm$0.3 & 62.2$\pm$1.5 & 60.6$\pm$1.2 & 53.5$\pm$1.5 & 43.7$\pm$0.4 & 25.3$\pm$0.1 & 22.5$\pm$1.0 & 35.4$\pm$1.7 & 26.9$\pm$0.7\\
        GeoReasoning & \textbf{48.6$\pm$0.3} & \textbf{62.8$\pm$1.3} & \textbf{61.4$\pm$1.4}  & \textbf{54.3$\pm$1.2} & \textbf{46.0$\pm$0.5} & \textbf{25.8$\pm$0.1} & \textbf{24.0$\pm$0.8} & \textbf{36.8$\pm$0.4} & \textbf{28.4$\pm$0.5}\\
        \bottomrule
    \end{tabular}}
    \label{tab3:compare}
    \vspace{-0.4cm}
\end{table*}

\begin{figure}[!t]
    \centering
    \begin{subfigure}[b]{0.45\linewidth} 
        \centering
        \includegraphics[width=\linewidth]{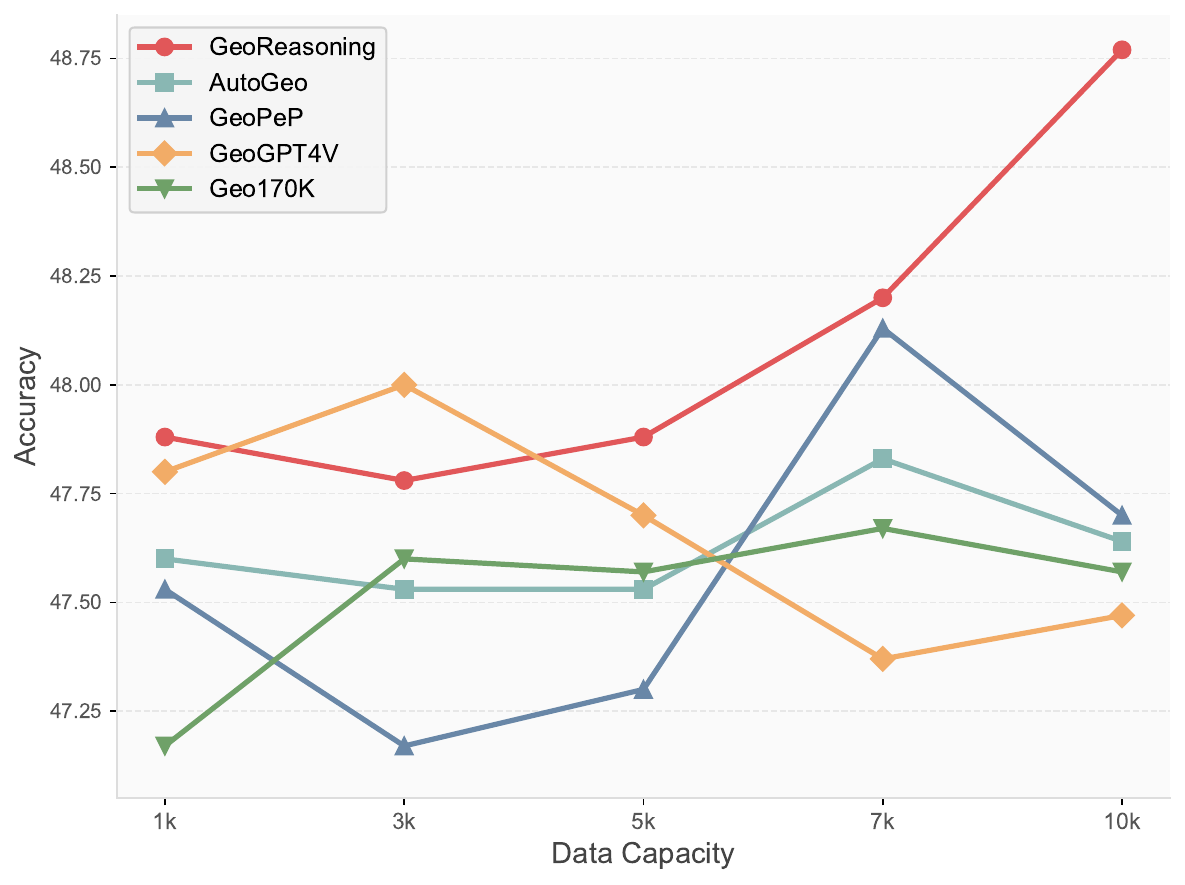}
        \caption{MathVista}
        \label{fig2-mathvista}
    \end{subfigure}%
    \begin{subfigure}[b]{0.45\linewidth}
        \centering
        \includegraphics[width=\linewidth]{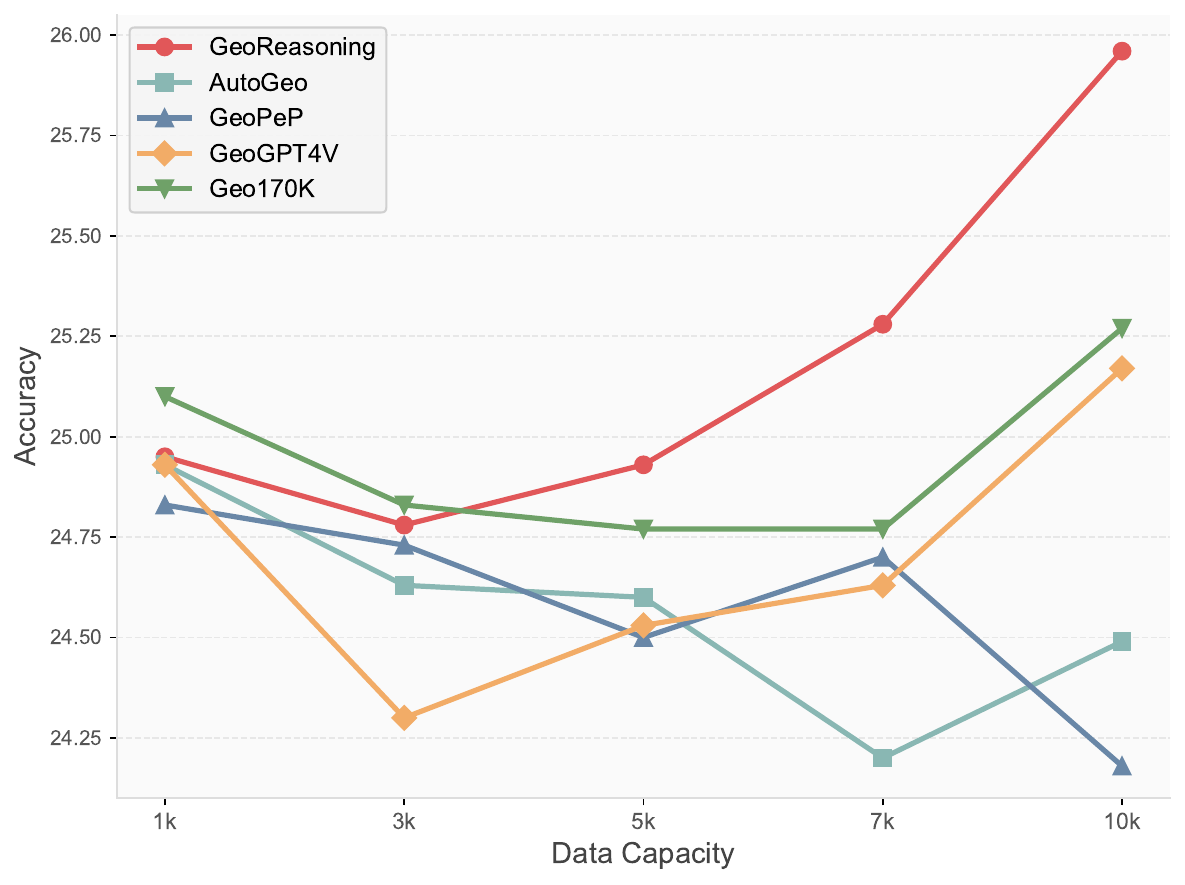}
        \caption{MathVerse}
        \label{fig2-mathverse}
    \end{subfigure}%
    \caption{\textbf{Better Scalability}. The accuracy of models fine-tuned on different capacities and mathematical datasets on downstream evaluation benchmarks: (a) MathVista. (b) MathVerse.}
    \label{fig2-mathvista_mathverse}
    \vspace{-0.4cm}
\end{figure}

GeoReasoning also demonstrates better scalability, as shown in Figure~\ref{fig2-mathvista_mathverse}. The model trained on GeoReasoning improves progressively when the dataset sizes increase. Moreover, GeoReasoning outperforms existing datasets by a non-trivial margin at 10K size, verifying the effectiveness of the proposed cross-modal alignment method. In this setting, subsets of different sizes are randomly sampled from the original dataset, with the baseline models trained on these subsets. 

\subsection{Out-of-Domain Performance of GeoReasoning-10K}
\label{sec:exp-gene}
Surprisingly, GeoReasoning-10K also demonstrates better out-of-domain generalization ability for non-geometric input images. Specifically, we evaluate the accuracy of the baseline models and the trained models on a commonly-used benchmark MMMU~\citep{mmmu}, as shown in Table~\ref{tab2:generalization}.

\begin{table}[!h]
\renewcommand{\baselinestretch}{1.0}
\renewcommand{\arraystretch}{1.1}
    \centering
    \caption{\textbf{Better Out-of-Domain Performance}. Accuracy of models evaluated on all subtasks of MMMU over 5 trials, where ``A\&D'', ``Busi'', ``Sci'', ``H\&M'', ``Human'', ``Tech'' are short for ``Art and Design'', ``Business'', ``Science'', ``Health and Medicine'', ``Humanities and Social Science'', ``Tech and Engineering'', respectively.}
    \label{tab2:generalization}
    \resizebox{\linewidth}{!}{
    \begin{tabular}{l|ccccccc}
        \toprule
         & \textbf{Overall} & \textbf{A\&D} & \textbf{Busi} & \textbf{Sci} & \textbf{H\&M} & \textbf{Human} & \textbf{Tech}  \\
        \midrule
        Base &    43.3$\pm$0.7&	57.8$\pm$4.0 &	44.1$\pm$0.6&	34.3$\pm$0.9&	46.8$\pm$2.2&	59.2$\pm$2.1 & 29.0$\pm$1.3\\
        AutoGeo & 43.5$\pm$0.5 & 59.3$\pm$1.4 & 43.3$\pm$1.1 & 34.9$\pm$1.3 & 47.4$\pm$1.1 & 58.9$\pm$1.5 & 30.7$\pm$2.6 \\
        GeoPeP & 43.7$\pm$0.9 & 59.2$\pm$1.1 & 40.4$\pm$1.4 & 34.0$\pm$1.7 & 45.1$\pm$0.9 & 59.6$\pm$1.0 & 32.6$\pm$0.7\\
        GeoGPT4V & 44.0$\pm$0.7 &\textbf{60.2$\pm$1.1} & 43.1$\pm$1.5 & 34.5$\pm$0.7 & 46.0$\pm$0.7 & 58.3$\pm$1.6 & 30.8$\pm$2.0\\
        Geo170K & 42.9$\pm$1.0 & 58.5$\pm$0.8 & 43.6$\pm$1.4 & 30.9$\pm$2.0 & 46.8$\pm$2.2 & 59.9$\pm$2.2 & 30.9$\pm$1.6 \\
        GeoReasoning & \textbf{44.9$\pm$0.7} & \textbf{60.2$\pm$2.0} & \textbf{44.5$\pm$2.5} & \textbf{36.0$\pm$2.0} & \underline{46.7$\pm$1.1} & \textbf{60.0$\pm$0.5} & \textbf{32.9$\pm$1.3}\\
        \bottomrule
    \end{tabular}
    }
\end{table}

It can be observed from Table~\ref{tab2:generalization} that the trained model outperforms the baseline on most of the domains, especially subtasks involving line art or drawing, such as Art \& Design, and Tech \& Engineering. These types of improvement can be mainly attributed to the proposed RLVR training process, whose reasoning reward forces the model to focus on the key element in the geometric image for problem-solving. This key-element capturing ability can help model generalize better to scenarios other than geometric problems.

\begin{figure}[!t]
    \centering
    \includegraphics[width=0.9\linewidth]{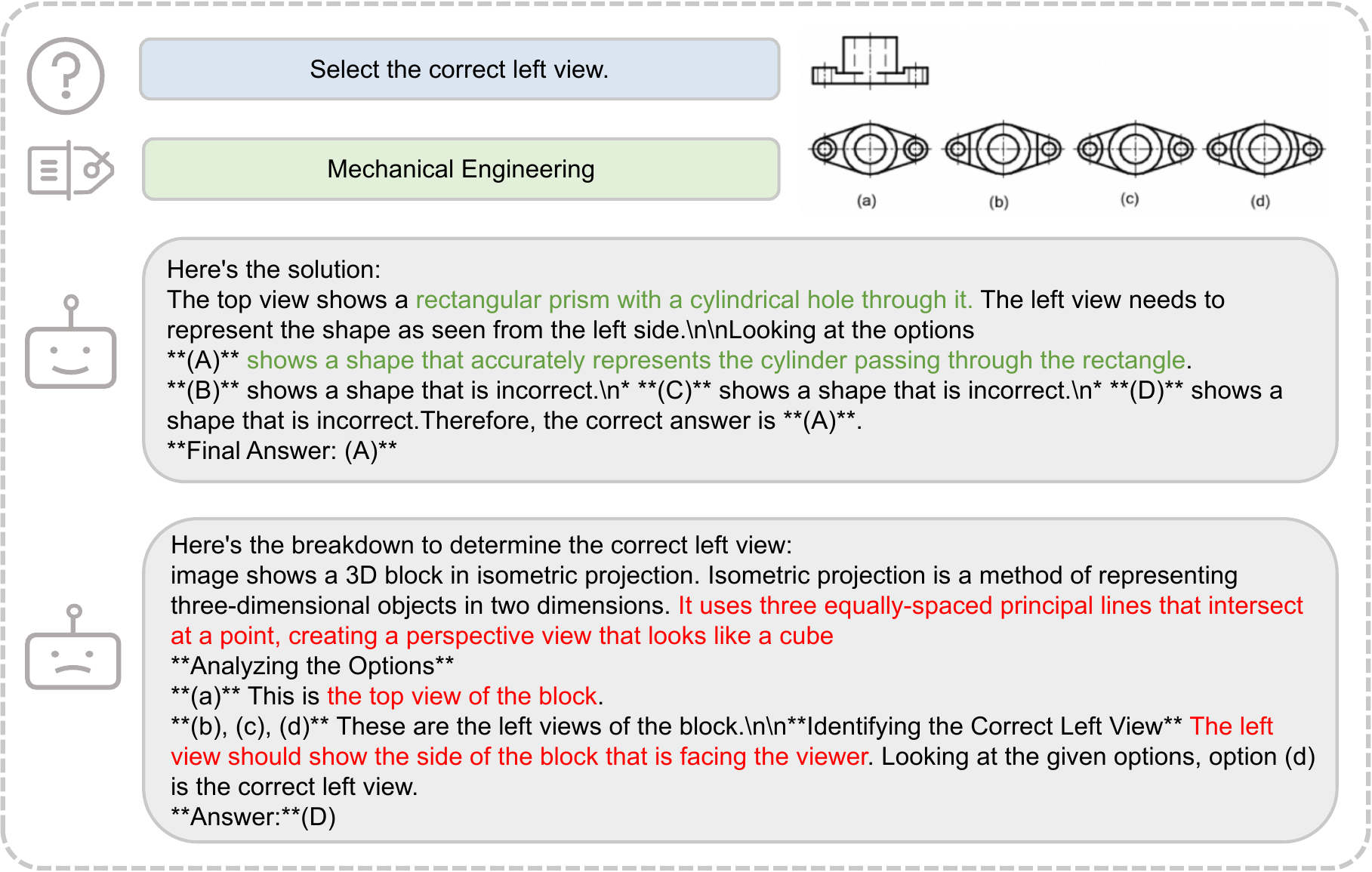}
    \caption{An engineering case, where the base model's answer is relatively general and the analysis of shape is not rigorous enough, while the model after training on GeoReasoning is more detailed and accurate in observing shape and has spatial reasoning ability.}
    \label{case_engineering}
\end{figure}

We also conduct qualitative analyses on representative examples from MathVista and MMMU to understand which types of reasoning abilities our method generalizes especially well. As shown in Figure~\ref{case_algebraic} and~\ref{case_physics}, GeoReasoning is in particular favorable for arithematic reasoning, line-art reasoning, and spatial reasoning. Additional examples are also available in Appendix~\ref{appendix:case}.

\subsection{Ablation Study}
\label{exp-1}
To understand the individual contribution of Cold-Start and RLVR phases, additional ablation studies are conducted. Specifically, we implement the Cold Start and RLVR pipelines on Gemma3-4B, generating refined models and datasets at each optimization stage. Models of various stages are then evaluated on MathVista and MathVerse, where the results are shown in Table~\ref{tab:raft}.
\begin{table}[!h]
\renewcommand{\baselinestretch}{1.0}
\renewcommand{\arraystretch}{1.1}
\setlength{\tabcolsep}{15pt}
    \centering
    \caption{Accuracy of Gemma3-4B models of various stages evaluated on MathVista and MathVerse.}
    \label{tab:raft}
    \begin{tabular}{l|ccc}
        \toprule
        & \textbf{MathVerse} & \textbf{MathVista}  \\
        \midrule
        Gemma3-4B & 25.2 & 46.2\\
        Gemma3-4B-Coldstart & 25.9 & 47.6\\
        Gemma3-4B-RAFT & 26.1 & 49.4\\
        Gemma3-4B-Coldstart-RAFT & \bf27.4 & \bf50.0\\
        \bottomrule
    \end{tabular}
\end{table}

It can be concluded that both cold-start and RLVR are effective. Furthermore, RLVR is helpful for both the base model and the one after cold-start, and cold-start is essential for enabling the model to acquire task-related capabilities, especially when the base model performs poorly in the given domain, such as MathVerse.

\section{Conclusion}
In this paper, we propose \textbf{Geo-Image-Textualization}, a novel reinforcement learning-based framework designed to symbolically synthesize high-quality, geometry-centered multimodal data. Leveraging this framework, we construct \textbf{GeoReasoning-10K}, a new dataset aimed at bridging the gap between visual and linguistic modalities in the geometry domain. Extensive experiments on the MathVista and MathVerse benchmarks demonstrate that our framework significantly enhances the cross-modal reasoning capabilities of MLLMs when fine-tuned on \textbf{GeoReasoning-10K}, with improvements generalizing to other domains, even with non-geometric input images. These results highlight the potential of rule-based symbolic synthesis for cross-modal and cross-domain learning, pointing to a promising direction for future research at the intersection of multimodal reasoning.



\bibliographystyle{unsrt}  
\bibliography{references}

\newpage
\appendix
\section{Question-Answer Pair Generation Prompt}
\label{app:qa}
The rule-and-LLM-based pipeline contains two stages. We first design a prompt that satisfies all the above conditions, using a relatively low temperature (0.2 in our experiments) to encourage the large model (Gemini 2.5 Flash) to generate initial questions based on the caption, while also labeling those that are inconsistent with the caption. For the inconsistent questions, we then switch to a different prompt, encouraging the model to incorporate additional information and formulate new questions accordingly, while increasing the temperature to 0.8. This process continues until a self-consistent question is generated for the first time.

The prompt of the first question generation stage is set as:
\begin{tcolorbox}[
    enhanced,
    breakable,
    width=\columnwidth,
    arc=3mm,
    boxrule=1pt,
    left=2mm,
    right=2mm,
    top=1mm,
    bottom=1mm,
    fontupper=\ttfamily,
    before skip=5mm,
    after skip=5mm,
    title=Prompt1, 
    coltitle=white, 
    colbacktitle=black, 
    fonttitle=\bfseries 
    ]
You are a helpful dataset processor. Please:

1. Generate a mathemetical question according to the given description of a geometric image with the following requirements:\\
    1.1 The problem should base on the given description, i.e., you must **NOT** generate problems that are unrelated to the given description. \\
    1.2 The problem difficulty should not be too low, such as determining some information in the description. \\
    1.3 It should also not be too hard, like introducing too much extra information, but anyway you can introduce some extra information to form a good geometric problem. \\
    1.4 You should **NOT** include or repeat any information in the description, and just contain the real question. For example, if the description says: `Line segment AB is present. The length of BA is 1.24.', then when you generate the question, you should not include the length of AB.\\
    1.5 If the question is inconsistent with the given description, the final answer should be `None'.\\
2. Answer the question you just provided, and express the final answer to two decimal places. The final answer should be in \textbackslash\textbackslash boxed\{\{answer\}\}.\\

Description: \\
\{description\}\\
Generated Question:\\
\{question\}\\
Generated Response:\\
\{response\}\\
Final Answer:\\
\textbackslash\textbackslash boxed\{\{answer\}\}
\end{tcolorbox}

\newpage
The prompt of the question re-generation stage is set as:
\begin{tcolorbox}[
    enhanced,
    breakable,
    width=\columnwidth,
    arc=3mm,
    boxrule=1pt,
    left=2mm,
    right=2mm,
    top=1mm,
    bottom=1mm,
    fontupper=\ttfamily,
    before skip=5mm,
    after skip=5mm,
    title=Prompt1, 
    coltitle=white, 
    colbacktitle=black, 
    fonttitle=\bfseries 
    ]
You are a helpful dataset processor. Please:
1. Generate a mathemetical question according to the given description of a geometric image with the following requirements:\\
    1.1 The problem should base on the given description, i.e., you must **NOT** generate problems that are unrelated to the given description. \\
    1.2 You can introduce some extra information to form a good geometric problem.\\ 
    1.3 If you find that it is hard to generate some difficult questions, just give a simple question. For example, when the lengths of all four sides of a quadrilateral are given, you can no longer assume it is a parallelogram or rectangle. In such cases, the problem may only allow for questions like asking for the perimeter, or determining the length of a segment when a certain point divides a side into an n-equal part, etc.\\
    1.4 You should **NOT** include or repeat any information in the description, and just contain the real question. For example, if the description says: `Line segment AB is present. The length of BA is 1.24.', then when you generate the question, you should not include the length of AB.\\
    1.5 If the question is inconsistent with the given description, the final answer should be `None'.\\
2. Answer the question you just provided, and express the final answer to two decimal places. The final answer should be in \textbackslash\textbackslash boxed\{\{answer\}\}.

Description: \\
\{description\}\\
Generated Question:\\
\{question\}\\
Generated Response:\\
\{response\}\\
Final Answer:\\
\textbackslash\textbackslash boxed\{\{answer\}\}
\end{tcolorbox}

\newpage
\section{Experimental Setup and Dataset Information}
\subsection{Experimental Setup}

The training and optimization pipeline contains two stages:
\begin{enumerate}
    \item \textbf{Cold-Start phase}: we train each base model on the initial GeoReasoning-10K dataset for 1 epoch using standard supervised fine-tuning (SFT). The peak learning rate is $10^{-5}$, with a cosine learning rate scheduler and a 0.03 warm-up ratio for linear warm-up.
    \item \textbf{RLVR phase}:
      we run RAFT~\citep{Dong2023RAFT} for 5 epochs, alternating between two sub-phases:
      \begin{itemize}
      \item 2.1) Caption Refinement: The model generates 8 candidate captions for each image, and the top-1 caption per image is selected based on a composite reward with reasoning reward weight of $\lambda_r = 0.7$ and caption reward weight of $1 - \lambda_r = 0.3$.
      \item 2.2) Model Retraining: Fine-tune the model on the selected dataset for 1 epoch using the same hyperparameters as the cold start phase.
    \end{itemize}
    
\end{enumerate}

To ensure consistency, we adopt the official evaluation codebases of both MathVerse and MathVista, using the GPT-4o-mini API to evaluate the performance of our MLLM. Specifically, following each benchmark's official setup, we use GPT-4o-mini to extract and assess the correctness of answers for MathVerse, and to extract answers for MathVista.

We evaluate MLLMs on MathVerse, MathVista, and MMMU using A100 by VLLM. We employ Gemma3-4B as our base model and fine-tune it on Georeasoning-10K using 4 L20 GPUs. The training process is distributed using torchrun with the DeepSpeed ZeRO-3 optimization strategy.

\subsection{Data Source}
GeoReasoning-10K dataset is generated through rule-based methods and further refined using the RAFT framework. The question-answer pairs are generated by Gemini 2.5-Flash with a specific prompt.

\subsection{License}

\begin{itemize}
  \item \textbf{GeoReasoning-10K} is released under the \href{https://opensource.org/licenses/MIT}{MIT License}.
  \item \textbf{MathVerse} and \textbf{MathVista} are evaluated using their official codebases, which are publicly available under the Apache 2.0 License and MIT License, respectively.
  \item Our use of the \textbf{GPT-4o-mini API} for evaluation complies with OpenAI’s API usage policies.
  \item All third-party datasets and models used in this work are under their respective licenses, and we ensure compliance with their terms of use.
\end{itemize}

\section{Case Studies}
\label{appendix:case}
\begin{figure}[!h]
    \centering
    \includegraphics[width=0.9\linewidth]{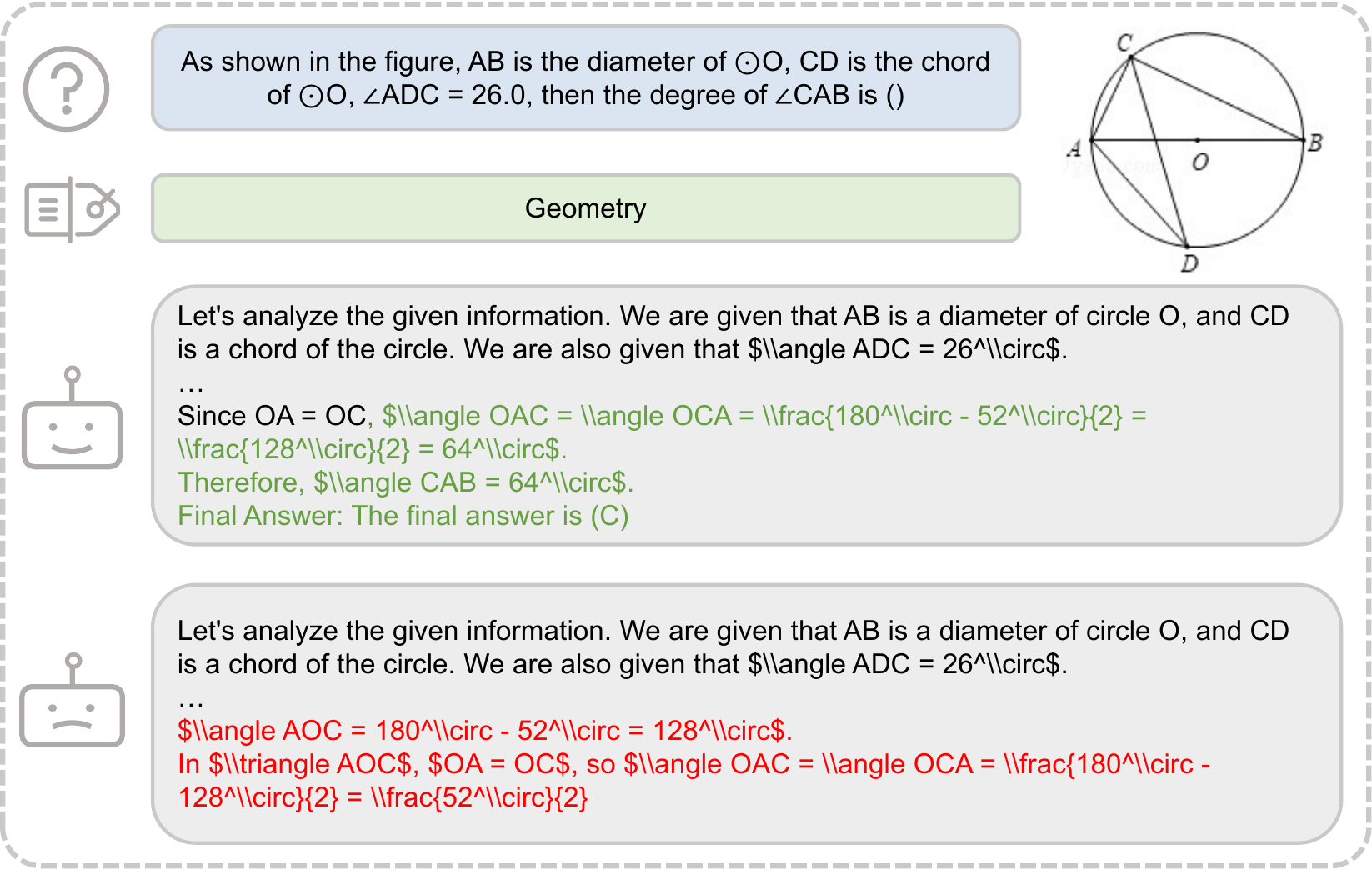}
    \caption{A geometric case.}
    \label{case_geometry}
\end{figure}

\begin{figure}[!h]
    \centering
    \includegraphics[width=0.9\linewidth]{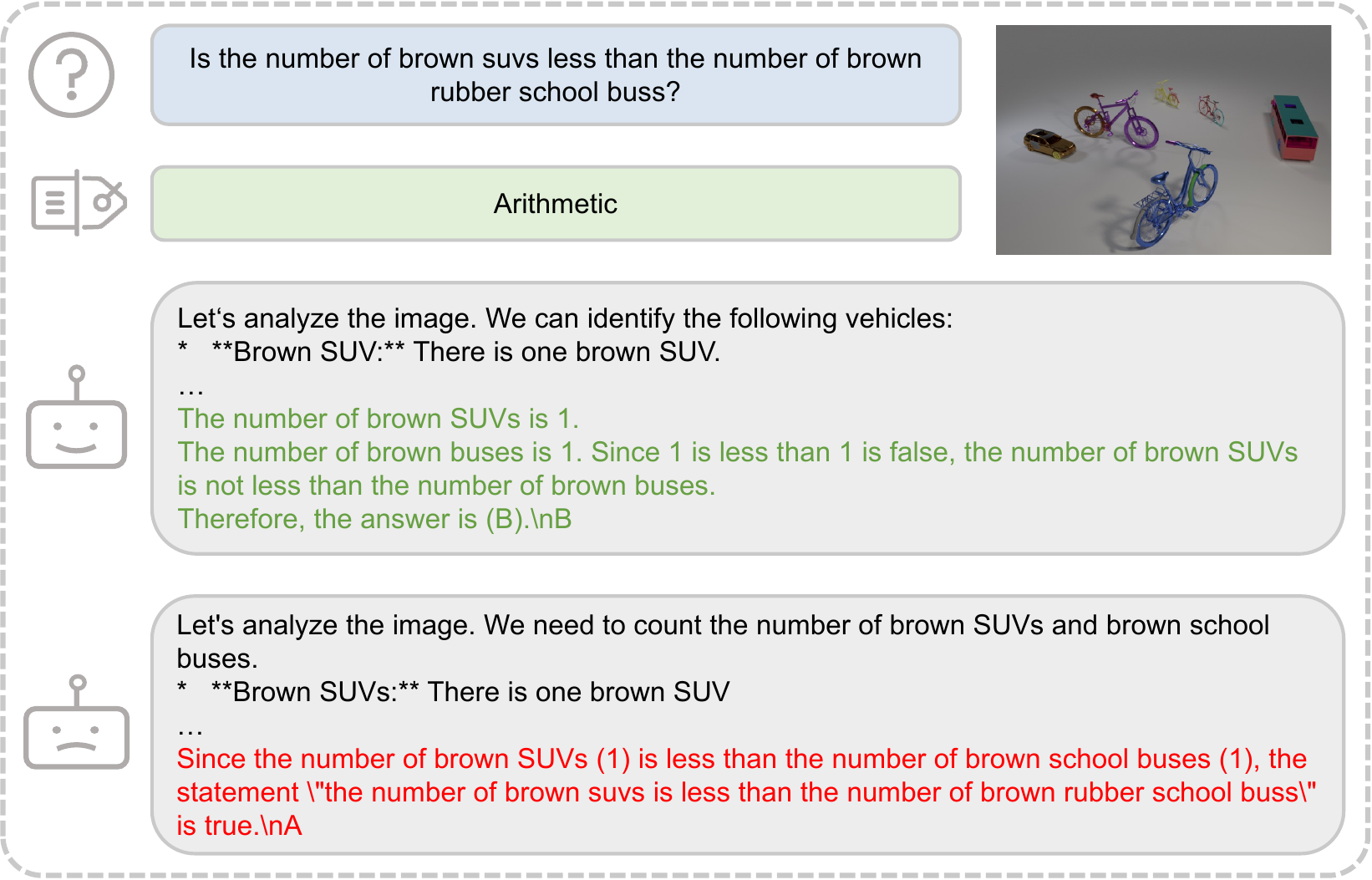}
    \caption{An arithmetic case.}
    \label{case_arithmetic}
\end{figure}

\begin{figure}[!h]
    \centering
    \includegraphics[width=0.9\linewidth]{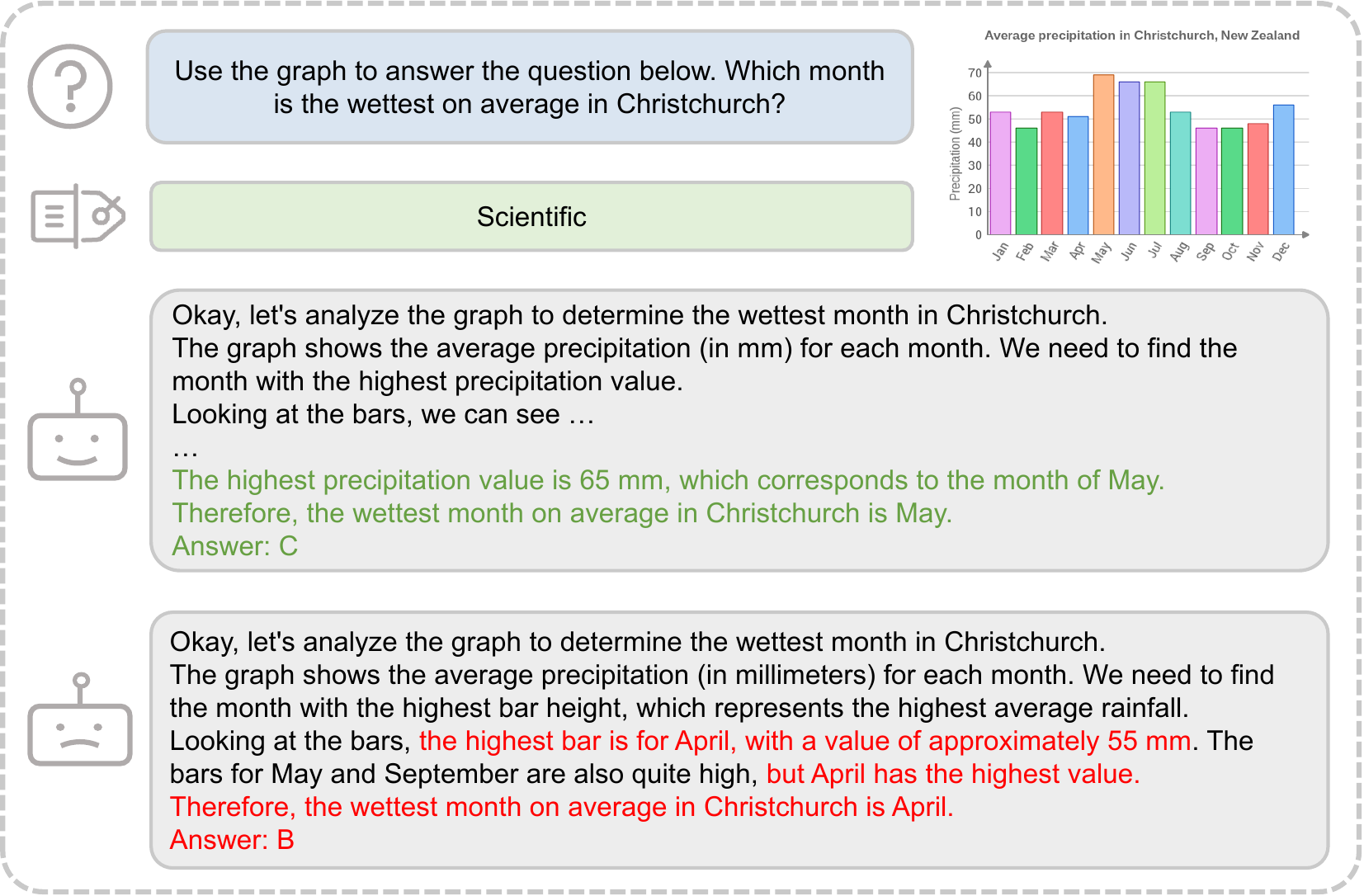}
    \caption{A numeric case.}
    \label{case_numeric}
\end{figure}

Additional mathematical cases that further demonstrate the generalization capability of our RAFT method are presented here. Figure~\ref{case_geometry}, Figure~\ref{case_arithmetic}, and Figure~\ref{case_numeric} correspond to the geometric, arithmetic, and numeric domains, respectively. 

\begin{figure}[!h]
    \centering
    \includegraphics[width=0.9\linewidth]{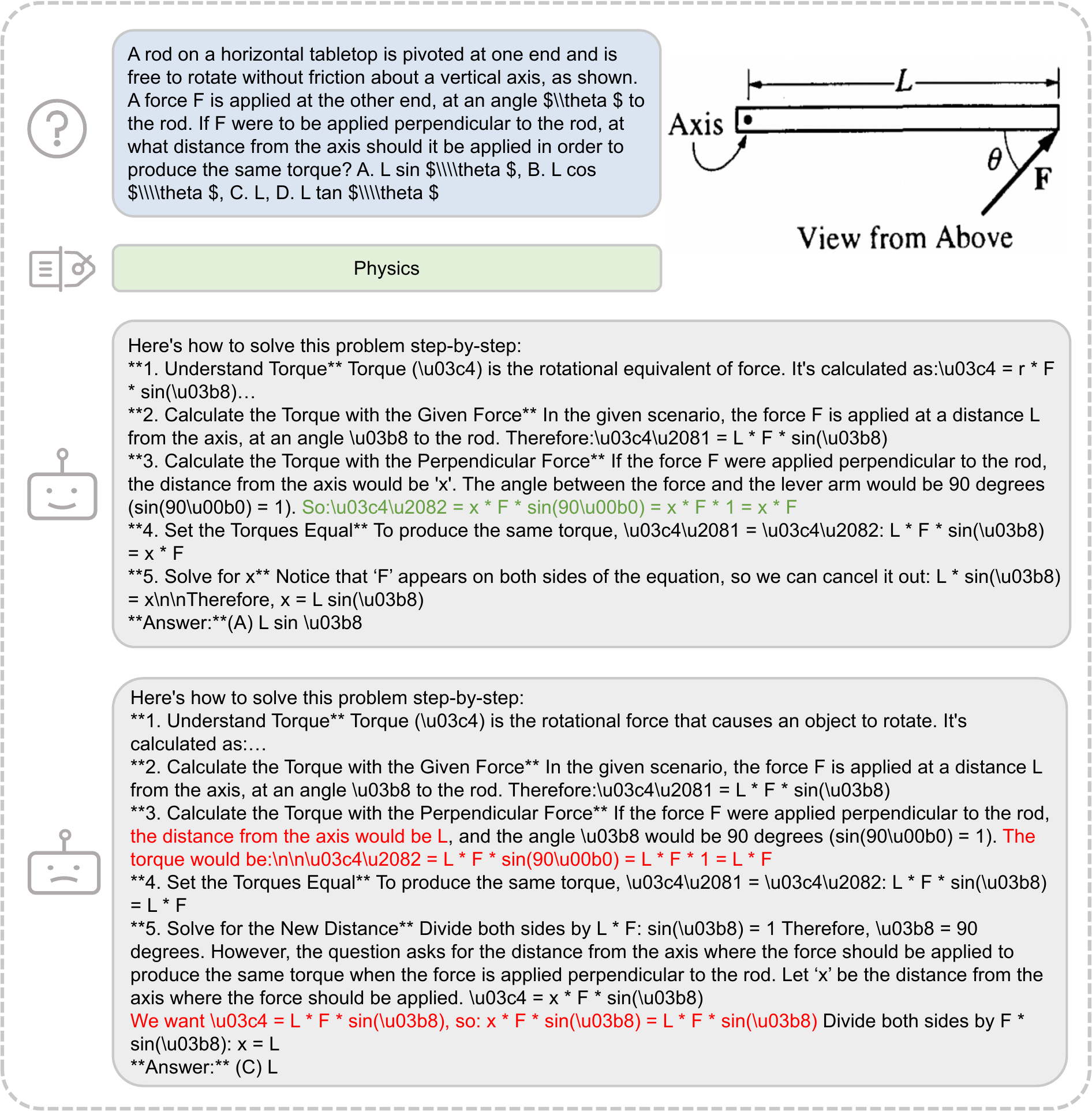}
    \caption{A physics case.}
    \label{case_physics}
\end{figure}

\begin{figure}[!h]
    \centering
    \includegraphics[width=0.9\linewidth]{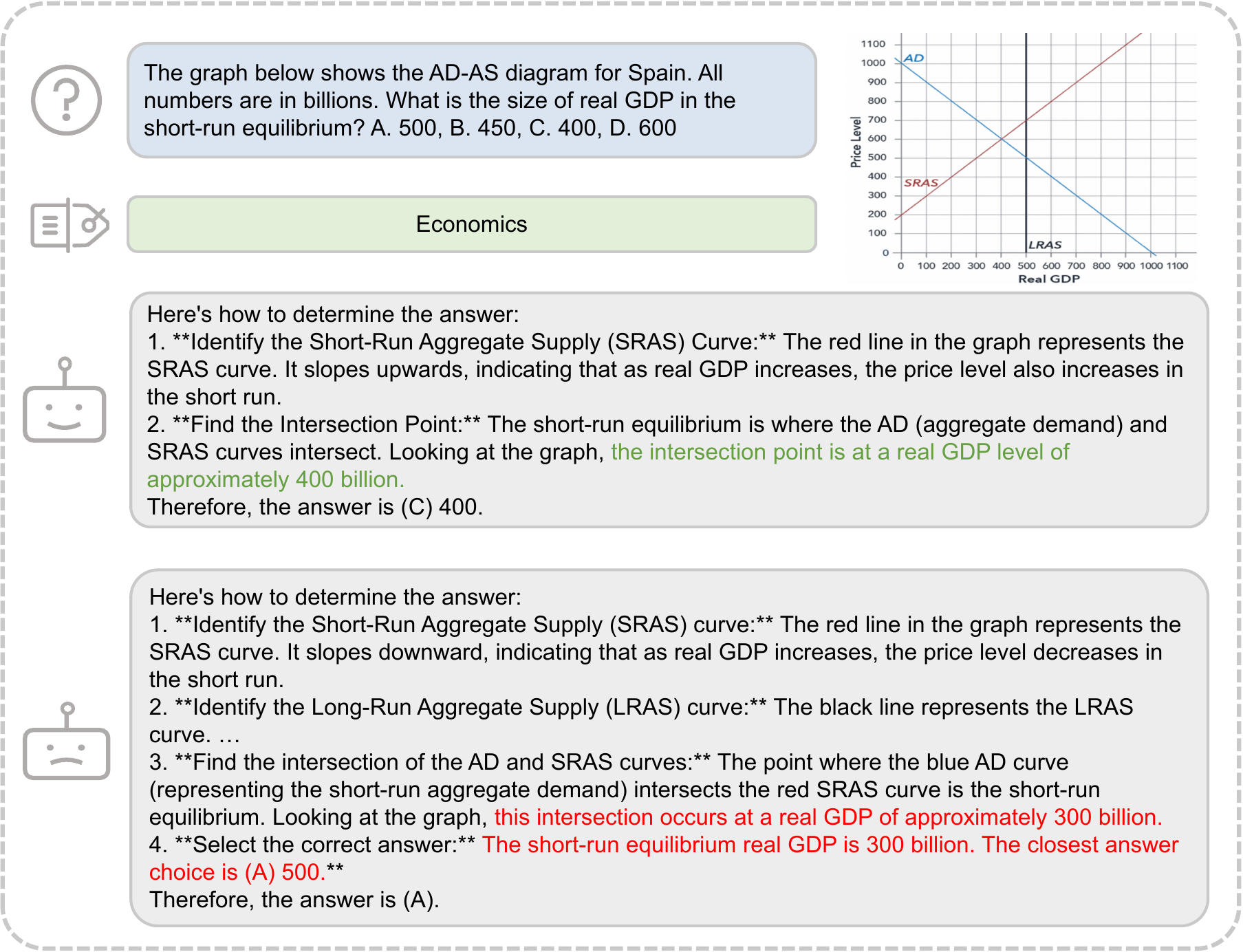}
    \caption{An economics case.}
    \label{case_economics}
\end{figure}

\clearpage
Figure~\ref{case_physics} and Figure~\ref{case_economics} exhibit non-mathematical cases of the physics and economics domains, respectively.

All these examples indicate that training on geometric caption tasks stimulates the reasoning capacity of models.

\section{Comparison to Other Datasets}
\label{app:sizes}
We test the model performance trained on total sizes of AutoGeo, GeoPeP, GeoGPT4V, Geo170K, GeoQA, MathVision, with the accuracy of MathVista and MathVerse shown in Table~\ref{tab3:sizes}:
\begin{table}[!h]
\renewcommand{\baselinestretch}{1.0}
\renewcommand{\arraystretch}{1.1}
\setlength{\tabcolsep}{15pt}
    \centering
    \caption{Performance of Gemma3-4B models trained on the total capacity of our dataset and counterpart datasets.}
    \label{tab3:sizes}
    \begin{tabular}{lc|ccc}
        \toprule
        & \textbf{Capacity} & \textbf{MathVerse} & \textbf{MathVista}  \\
        \midrule
        Geo170K & 117k & 22.0 & 46.8\\
        GeoPeP & 100k & 22.7 & 47.1\\
        GeoGPT4V & 23k & 24.6 & 46.0\\
        AutoGeo & 100k & 24.7 & 46.1\\
        MathVision & 3k & 24.7 & 46.9\\
        GeoQA & 5k & 24.9 & 46.0\\
        GeoReasoning & 10k & \bf25.8 & \bf48.6\\
        \bottomrule
    \end{tabular}
\end{table}


\newpage
\section{Additional Ablation Studies}
\label{sec:ablation}
This section serves as a complement of Section \ref{exp-1}, exhibit ing ablation studies on various domains, and hyperparameters of the reward function.

\subsection{Ablation Study on Various Domains}
We record the skills across diverse domains like geometry and arithmetic on various training stages in Table~\ref{tab1-mathvista} and Table~\ref{tab1-mathverse}:
\begin{table}[!h]
\renewcommand{\baselinestretch}{1.0}
\renewcommand{\arraystretch}{1.1}
\setlength{\tabcolsep}{8pt}
    \centering
    \caption{Accuracy of Gemma3-4B models at various stages tested on MathVista}
    \label{tab1-mathvista}
    \begin{tabular}{l|ccccccc}
        \toprule
        & \textbf{baseline} & \textbf{cold-start} & \textbf{raft-1} & \textbf{raft-2} & \textbf{raft-3} & \textbf{raft-4} & \textbf{raft-5}   \\
        \midrule
        all&	46.2&	47.6&	48.7&	48.1&	49.2&	49.0&	\bf 50.0\\
        geometry&	60.7&	62.3&	63.2&	64.0&	63.6&	60.3&	\bf 64.0\\
        arithmetic&	42.5&	45.0&	44.8&	45.3&	45.9&	\bf 47.6&	46.5\\
        algebraic&	59.1&	60.5&	62.3&	62.3&	62.3&	59.1&	\bf 63.3\\
        numeric&	26.4&	31.9&	29.9&	31.3&	31.3& 31.9&	\bf 31.9\\
        \bottomrule
    \end{tabular}
\end{table}

\begin{table}[!h]
\renewcommand{\baselinestretch}{1.0}
\renewcommand{\arraystretch}{1.1}
\setlength{\tabcolsep}{8pt}
    \centering
    \caption{Accuracy of Gemma3-4B models at various stages tested on MathVerse}
    \label{tab1-mathverse}
    \begin{tabular}{l|ccccccc}
        \toprule
        & \textbf{baseline} & \textbf{cold-start} & \textbf{raft-1} & \textbf{raft-2} & \textbf{raft-3} & \textbf{raft-4} & \textbf{raft-5}   \\
        \midrule
        all&    25.2&	25.9&	25.7&	25.8&	25.5&	26.5 & \bf27.4\\
        text dominant&  32.0&	35.5&	35.1&	35.2&	35.1&  36.5 & \bf36.5\\
        text lite&  25.9&	27.4&	28.2&	\bf28.5&	27.4&	26.6 & 26.3\\
        vision intensive&   24.0&	24.8&	24.4&	24.4&	23.1&	26.1 & \bf26.5\\
        \bottomrule
    \end{tabular}
\end{table}
It can be observed from Table~\ref{tab1-mathvista} and Table~\ref{tab1-mathverse} that the model after RAFT stages outperforms the base model across all domains. Specifically, the model achieves significant performance improvements across the arithmetic, algebraic, and numeric domains, with respective gains of 5.1\%, 4.2\%, and 5.5\%. These results demonstrate the effectiveness of our approach in enhancing model performance as well as its generalization capability across different domains.

\subsection{Ablation Study on Hyperparameters}
We evaluated the RAFTed models with various hyperparameters on MathVista and MathVerse, as shown in Table~\ref{tab:param}:
\begin{table}[!h]
\renewcommand{\baselinestretch}{1.0}
\renewcommand{\arraystretch}{1.1}
\setlength{\tabcolsep}{15pt}
    \centering
    \caption{Accuracy of RAFTed models with various hyperparameters evaluated on MathVista and MathVerse, where $\lambda_r$ stands for the weight of reasoning reward.}
    \label{tab:param}
    \begin{tabular}{l|ccc}
        \toprule
        & \textbf{MathVista} & \textbf{MathVerse}  \\
        \midrule
        $\lambda_r$ = 1 & 49.8 & \bf27.5\\
        $\lambda_r$=0.7 & \bf50.0 & 27.4\\
        $\lambda_r$=0 & 48.9 & 27.5\\
        \bottomrule
    \end{tabular}
\end{table}

As shown in Table~\ref{tab:param}, the reasoning reward plays a more important role in MathVista than MathVerse, indicating that the gain of genelization comes more from the helpness in solving the question other than comparison with captions.

In addition, it is observed in the result that the performance is not very sensitive to the selection of this hyperparameter, indicating the robustness of our RAFT method.

\section{Broader Impacts}\label{appendix:broader_impact}
The provided dataset pipeline and the generated dataset contribute to enhancing the generalizable reasoning abilities of multimodal large language models (MLLMs). In narrow domains, they are particularly effective for improving the geometric problem-solving capabilities of MLLMs, while in broader domains, they support the development of mathematical reasoning skills applicable to everyday scenarios. As the dataset is limited to geometric mathematical problems, it is considered safe for release and is unlikely to pose direct negative social impacts.
\end{document}